\documentclass{article}

\usepackage{arxiv}

\usepackage[utf8]{inputenc} 
\usepackage[T1]{fontenc}    
\usepackage{hyperref}       
\usepackage{url}            
\usepackage{booktabs}       
\usepackage{amsfonts}       
\usepackage{nicefrac}       
\usepackage{microtype}      
\usepackage{lipsum}		
\usepackage{graphicx}

\usepackage{doi}

\title{A template for the \emph{arxiv} style}

\usepackage{epstopdf}
\usepackage{amsmath,amssymb,amsfonts}
\usepackage{graphicx}
\usepackage{textcomp}
\usepackage{amssymb, amsmath}
\usepackage[hang,small,bf]{caption}
\usepackage[subrefformat=parens]{subcaption}
\usepackage{algorithm}
\usepackage{algpseudocode}
\usepackage{makecell}

\def\BibTeX{{\rm B\kern-.05em{\sc i\kern-.025em b}\kern-.08em
    T\kern-.1667em\lower.7ex\hbox{E}\kern-.125emX}}

\usepackage[backend=biber,style=ieee]{biblatex}
\usepackage[flushleft]{threeparttable}
\bibliography{cas-refs.bib}

\begin{document}
\title{Data Collaboration Analysis applied to Compound Datasets and the Introduction of Projection data to Non-IID settings}

\author{
 Akihiro Mizoguchi \\
  University of Tsukuba \\
  \texttt{s2030212@u.tsukuba.ac.jp} \\
   \And
 Anna Bogdanova \\
  University of Tsukuba \\
  \texttt{bogdanova.anna.fw@u.tsukuba.ac.jp} \\
  \And
 Akira Imakura \\
  University of Tsukuba \\
  \texttt{imakura@cs.tsukuba.ac.jp} \\
  \And
 Tetsuya Sakurai \\
  University of Tsukuba \\
  \texttt{sakurai@cs.tsukuba.ac.jp} \\
}

\thanks{This work was supported in part by the New Energy and Industrial Technology Development Organization (NEDO) and the Japan Society for the Promotion of Science (JSPS) Grants-in-Aid for Scientific Research [grant numbers JP21H03451, JP22H00895, and JP22K19767].}

\maketitle

\begin{abstract}
Given the time and expense associated with bringing a drug to market, numerous studies have been conducted to predict the properties of compounds based on their structure using machine learning. Federated learning has been applied to compound datasets to increase their prediction accuracy while safeguarding potentially proprietary information. However, federated learning is encumbered by low accuracy in not identically and independently distributed (non-IID) settings, i.e., data partitioning has a large label bias, and is considered unsuitable for compound datasets, which tend to have large label bias. To address this limitation, we utilized an alternative method of distributed machine learning to chemical compound data from open sources, called data collaboration analysis (DC). 
We also proposed data collaboration analysis using projection data (DCPd), which is an improved method that utilizes auxiliary PubChem data. This improves the quality of individual user-side data transformations for the projection data for the creation of intermediate representations.
The classification accuracy, i.e., area under the curve in the receiver operating characteristic curve (ROC-AUC) and AUC in the precision-recall curve (PR-AUC), of federated averaging (FedAvg), DC, and DCPd was compared for five compound datasets. We determined that the machine learning performance for non-IID settings was in the order of DCPd, DC, and FedAvg, although they were almost the same in identically and independently distributed (IID) settings. Moreover, the results showed that compared to other methods, DCPd exhibited a negligible decline in classification accuracy in experiments with different degrees of label bias. Thus, DCPd can address the low performance in non-IID settings, which is one of the challenges of federated learning.
\end{abstract}

\section{Introduction}
\label{sec:introduction}
It is time-consuming and expensive to research, develop, and launch a drug. According to past surveys, the mean cost ranges from \$314 million to \$2.8 billion \cite{wouters2020estimated} and it takes 12–15 years \cite{hughes2011principles} to bring a drug to market. This is because of the need to measure various physical properties, affinity with target molecules, toxicity, and pharmacokinetics of candidate compounds. In addition, non-clinical and clinical studies using animals must be conducted. As such, several studies have been conducted to predict the properties of compounds based on their structures using machine learning methods. For example, there is a set of compound datasets of toxicity called Tox21 \cite{mayr2016deeptox}, and these datasets have been used by investigators to improve the accuracy of toxicity predictions. In general, it is known that the performance of machine learning increases as the number of training data increases. Therefore, it is expected that the accuracy of machine learning will be improved if data from multiple pharmaceutical companies are utilized for integrated analysis. However, it is challenging to share raw data on compound structures among pharmaceutical companies without inadvertently disclosing proprietary information.

In order to perform machine learning on data from multiple sources without aggregating raw data, federated learning was proposed by Google in 2017 \cite{konevcny2016federated}. This approach has been applied in various contexts, such as predicting emojis in Google's software keyboard, Gboard \cite{hard2018federated}, and diagnosing skin diseases from images \cite{hossen2022federated}. As such, 10 European pharmaceutical companies have gathered in a consortium called Machine Learning Ledger Orchestration for Drug Discovery (MELLODDY), to improve the performance of machine learning using compound data from multiple pharmaceutical companies, without sharing raw data, using federated learning \cite{burki2019pharma}. However, the compound libraries of pharmaceutical companies vary widely from one company to another. Therefore, the compound datasets of each pharmaceutical company often have large label biases, for which the label distributions of compounds across pharmaceutical companies are significantly different. On the other hand, federated learning is known to exhibit poor performance in this case \cite{zhao2018federated}. It is one of the non-IID (not independently and identically distributed) settings. In general, datasets that contain partitioning with different label distributions or sample sizes are called non-IID datasets. However, non-IID datasets in this study are defined as datasets with different label distributions. 

Numerous studies have been conducted on federated learning to achieve acceptable machine learning performance using non-IID datasets \cite{zhu2021federated}. They include data sharing \cite{zhao2018federated}, data augmentation \cite{tanner1987calculation}, and knowledge distillation \cite{hinton2015distilling}. Data sharing facilitates high machine learning performance in non-IID settings by allowing each user to share only a small percentage of the total data. Using a data sharing method, it was shown that 5\% globally shared data increased the accuracy by 30\% in the non-IID split CIFAR-10 dataset. Data augmentation methods include the mix-up method \cite{yoon2021fedmix} and federated generative adversarial network (GAN) data augmentation \cite{li2022federated}. These methods reduce label imbalance by augmenting data, but they require the upload of some local data to a server, which leads to a compromise of privacy. Knowledge distillation is a machine learning technique in which the output of the teacher's model is imitated by the student model. Federated Learning
via Model Distillation (FedMD) \cite{li2019fedmd} and Federated Distillation Fusion (FedDF) \cite{lin2020ensemble} are examples of methods that apply knowledge distillation to non-IID federated learning. In FedMD, the knowledge from a public dataset can be shared by each client. Initially, the clients' models are trained on a portion of privacy-sensitive labeled public data, then subsequently trained on private data. In FedDF, after training with federated averaging, the parameters of the aggregated model are updated using unlabelled public data, which improves the accuracy of machine learning in the non-IID settings compared to FedAvg. Nevertheless, federated learning methods that use knowledge distillation can still be improved because their accuracy in non-IID settings is still much lower compared to that of IID settings \cite{lin2020ensemble}.

For machine learning involving multiple organizations without the sharing of raw data in a similar manner to federated learning, data collaboration analysis was first introduced in 2020 \cite{imakura2020data}. Unlike federated learning, which iteratively aggregates model updates from the users, data collaboration analysis relies on irreversible user-side transformations of data. These are known as intermediate representations, which are shared with the server and further projected to a common latent space for model training. The task of unifying the projection is achieved using a shared dataset, known as the anchor data. For more details on the method, the original paper should be reviewed \cite{imakura2020data}. In previous studies, data collaboration analysis was applied to tabular and image data \cite{imakura2021collaborative}, \cite{imakura2021interpretable}, \cite{imakura2022dc}, \cite{imakura2023non}, and a comparison with federated averaging was conducted for the image data in the IID setting \cite{bogdanova2020federated}.

In the present investigation, we extend the study of the applicability of data collaboration analysis to distributed chemical compound data. In addition, we present a novel solution involving data collaboration analysis using projection data (DCPd). This facilitates improved performance in non-IID datasets containing distributed data for different label distributions across users by introducing ``the projection data" from public databases. We then evaluate the proposed method by comparing it to federated averaging (FedAvg) and conventional data collaboration analysis (DC).

The major contributions of this study are as follows:

\begin{itemize}
      \item We show that data collaboration analysis can be applied to compound data using unlabelled public data as the anchor data.
      \item We show that data collaboration analysis facilitates improved classification accuracy compared to FedAvg, which is the main algorithm for federated learning in non-IID cases.
      \item We propose a novel method, DCPd, by introducing the projection data from public databases into data collaboration analysis, and establish that the classification accuracy is higher compared to that of FedAvg or DC for non-IID cases.
\end{itemize}

\section{Methodology}
\subsection{Federated averaging (FedAvg)}
This section provides an overview of federated averaging (FedAvg) \cite{mcmahan2017communication}--a fundamental algorithm used in federated learning \cite{konevcny2016federated}. Algorithm 1 presents the pseudo code for federated averaging using variables $n$ (number of clients or users), $s_i$ (sample size of the $i$th client or user), and $m$ (number of features), as well as matrices $X_i \in R^{s_{i} \times {m}}$, representing the training dataset of the $i$th client or user, and $Y_i \in R^{s_i}$, representing the ground truth for the training dataset. Additionally, the test dataset $X^{test}$ and predicted ground truth $Y^{test}$ are defined, along with the global model $\theta_{t}$ for the $t$th communication round and the local model $\theta_{t^k}$ for the $k$th client in the office $t$th communication round. The parameters $T$ (total communication round), $d$ (number of clients participating in learning in each round), $E$ (number of learning epochs for each client), $B$ (minibatch size), $\lambda$ (learning rate), and $L$ (loss function) are also introduced. 

The process begins with the server distributing its global model $\theta_{0}$ to a selected number of clients ($d$), each of which modifies their local model to be $\theta_{1^k}$ while learning $E$ times using their own data. The subsequent global model $\theta_{1}$ is generated by merging the local models from the participating clients. This process is repeated $T$ times to produce the final global model $\theta_{T}$, which is then applied to the test dataset $X^{test}$ to obtain $Y^{test}$.

\begin{algorithm}
  \caption{Federated Averaging (FedAvg) \cite{mcmahan2017communication}}
  \label{alg:hoge}
  \begin{algorithmic}[1]
    \Require{$X_{i} \in R^{s_{i}\times{m}}, Y_{i} \in R^{s_{i}}$}
    \Ensure{$Y^{test}=\theta_{T}(X^{test})$}
    \State \textbf{SERVER SIDE:}
    \State {Initialize $\theta_{0}$}
    \For{each round ${t} = 1 \, \ldots \, T$}
        \State {$S_{n} \leftarrow $ {set of} $n$ {clients}}
        \For{each client ${k} \in S_{n}$ in parallel}
            \State {$\theta^{k}_{t} \leftarrow $ ClientUpdate(${\theta_{t-1}}$)}
        \EndFor
        \State{$\theta_{t} \leftarrow \sum^{n}_{k=1}\frac{s_k}{s}\theta^{k}_{t},$ where $s=\sum^{n}_{k=1}s_k$}
    \EndFor
    \State \textbf{Client $k$ update:}
    \For{each local epoch  $e = 1 \, \ldots \, E$}
        \For{each minibatch $b\in{B}$}
            \State {$\theta \leftarrow \theta - \lambda\nabla{L}(\theta,b)$}    
        \EndFor
    \EndFor
  \end{algorithmic}
\end{algorithm}

\subsection{Conventional data collaboration analysis (DC)}
\label{sec:sample1}
In this section, the conventional data collaboration analysis is described (Algorithm 2). Through the use of dimensionality reduction methods, distributed datasets from multiple studies are transformed into intermediate representations and then collected to a server. Subsequently, these datasets are converted into collaborative representations through the application of supervised and unsupervised learning methods, such as principal component analysis (PCA) \cite{pearson1901liii}, partial least squares-discriminant analysis (PLS-DA) \cite{barker2003partial}, and random forest \cite{breiman2001random}.

Anchor data $X^{anc} \in R^{a \times m}$ that can be shared across all organizations is prepared, where $a$ is the number of anchor data. To protect privacy, the original data is converted into intermediate representations, $\tilde{X}_{i}=f_{i}(X_{i}), \tilde{X}_{i}^{anc}=f_{i}(X^{anc})$, using dimensionality reduction algorithms such as PCA \cite{pearson1901liii}, locality preserving projection \cite{he2003locality}, and uniform manifold approximation and projection for dimension reduction \cite{mcinnes2018umap}. Singular Value Decomposition (SVD) is then used to obtain $U_{1}$ from $\tilde{X}_{i}^{anc}$, and $G_{i}$ is computed such that $G_{i} = (\tilde{X}_{i}^{anc})^{\dag} U_{1}$. This $G_{i}$ satisfies the equations $\hat{X}_{i}^{anc}=\hat{X}_{i}^{anc}{G_{i}}, \hat{X}_{i}^{anc} \approx{X}_{j}^{anc}$, and $\hat{X}_{i}= \tilde{X}_{i}{G_{i}}$ is computed. $\hat{X}$ and $Y$ are then set such that $\hat{X}=[\hat{X}_{1}^{T},\hat{X}_{2}^{T},...,\hat{X}_{n}^{T}]^{T}, Y=[Y_{1}^{T},Y_{2}^{T},...,Y_{n}^{T}]^{T}$, where $\hat{X}_{i}^{T}$ is the collaboration representation of the $i$th user’s training dataset. A model $h$ is constructed using $\hat{X}$ as the training set and $Y$ as the ground truth, such that $Y \approx h(X)$. Finally, the objective variables of test data are predicted using the model $h$.

\renewcommand{\algorithmicrequire}{\textbf{Input:}}
\renewcommand{\algorithmicensure}{\textbf{Output:}}

\begin{algorithm}
  \caption{Data Collaboration analysis (DC) \cite{imakura2020data}}
  \label{alg:hoge}
  \begin{algorithmic}[1]
    \Require{$X_{i} \in R^{s_{i}\times{m}}, Y_{i} \in R^{s_{i}}$}
    \Ensure{$Y^{test}=h((f_{i}(X^{test}))G_{i})$}
    \State \textbf{USER SIDE:}
    \State \{Step 0. Preparation of Anchor data\}
    \State {Prepare $X^{anc} \in R^{a \times m}$}
    \State {Share $X^{anc}$ across all users}
    \State \{Step 1. Construction of Intermediate representation\}
\For{$i = 1,2,...,n$}
    \State {Compute $f_{i}$ as dimensionality reduction projection of $X_{i}$}
    \State {Construct $\tilde{X}_{i}=f_{i}(X_{i}), 
\tilde{X}_{i}^{anc}=f_{i}(X^{anc})$}
\EndFor
    \State {Centralize $\tilde{X}_{i}, \tilde{X}_{i}^{anc}$ and $Y_{i}$ for each $i$}
    \State \textbf{SERVER SIDE:}
    \State \{Step 2. Construction of Collaboration representation\}
    \State {Apply SVD to $\tilde{X}^{anc}=[\tilde{X}_{1}^{anc}, \tilde{X}_{2}^{anc},...,\tilde{X}_{n}^{anc}]$ $i.e.$  $\tilde{X}^{anc}=[U_{1},U_{2}]\begin{bmatrix}
\sum_{1} &  \\
 & \sum_{2} \\
\end{bmatrix}
\begin{bmatrix}
V_{1}^{T} \\
V_{2}^{T} \\
\end{bmatrix}$}
\For{$i = 1,2,..., n$}
    \State {Compute $G_{i}=(\tilde{X}_{i}^{anc})^{\dag}U_{1}$}
    \State {Compute $\hat{X}_{i}=\tilde{X}_{i}G_{i}$}
\EndFor
    \State {Set $\hat{X}=[\hat{X}_{1}^{T},\hat{X}_{2}^{T},...,\hat{X}_{n}^{T}]^{T}, Y=[Y_{1}^{T},Y_{2}^{T},...,Y_{n}^{T}]^{T}$}
    \State \{Step 3. Machine learning with Collaboration representation\}
    \State {Build a model $h$ by a machine learning method utilizing $\hat{X}$ and $Y$ as training data and ground truth, respectively $i.e.$ $Y \approx h(\hat{X})$}
    \State {Obtain $Y^{test}=h((f_{i}(X^{test}))G_{i})$}
  \end{algorithmic}
\end{algorithm}

\subsection{Basic idea of this study}
\label{sec:sample1}
When machine learning methods are applied to predict the properties of compounds based on their structures, methods that exploit feature extraction based on molecular fingerprints \cite{myint2012molecular}, the application of natural language processing using simplified molecular input line entry system (SMILES) \cite{karpov2020transformer}, and graph convolutional networks based on the application of structural equations as graph structures \cite{hung2021qsar} are often utilized. However, data collaboration analysis has not been applied to natural language processing or graph convolutional networks. Therefore, a method based on feature extraction using Morgan fingerprint \cite{rogers2010extended}, an example of a molecular fingerprint, is investigated in this study.

In this work, Morgan fingerprints with 2048 bits are used. They have a discrete distribution of 0 or 1, and a special distribution such that the number of 1s is negligibly small, and it is considered that existing anchor data creation methods may not be successful in data collaboration analysis. However, there are databases of chemical compounds such as PubChem \cite{wang2009pubchem}, ChemSpider \cite{pence2010chemspider}, and ChEMBL \cite{gaulton2012chembl}. In particular, we can extract SMILES format data using a Python library called "PubChemPy" (https: //pypi.python.org/pypi/PubChemPy), which can be used as the anchor data in data collaboration analysis.

In addition, it may not be possible to obtain an appropriate projection when dimensional reduction is performed to create intermediate representations in data collaboration analysis in cases where the distribution of the labels in each user's sample is biased. Therefore, we hypothesize that using each user's data and the compound data from PubChem to perform dimensional reduction and the application of the resulting projection to create intermediate representations will lead to improved performance compared to previous data collaboration analysis.

\subsection{Creation of anchor data from PubChem}
\label{sec:sample1}
In this study, anchor data was extracted by randomly selecting compounds from PubChem. This is different from previous data collaboration analysis methods, which used randomly generated anchor data \cite{imakura2020data}. 
This expectation is due to the anticipated improvement in performance that can be achieved by using data with a distribution similar to raw data as anchor data \cite{imakura2023another}. Therefore, it is expected that utilizing open data that closely resembles real data will lead to performance enhancement.
Specifically, we first extracted compounds with compound identifiers (CIDs) ranging from 1 to 12,000 in SMILES format using PubChemPy. The data were then converted into 2048-dimensional vector data using Morgan fingerprints. We then randomly selected an appropriate number and used them as anchor data for data collaboration analysis.

\subsection{Introduction of projection data from PubChem}
\label{sec:sample1}
As shown in step 1 of Fig. 1, in conventional data collaboration analysis methods, the projection from raw data to intermediate presentation is calculated using only each user's raw data. It is then applied to each user's raw data and common anchor data, and converted to an intermediate representation. However, it is thought that the appropriate projection cannot be calculated from each user's raw data only when the labels of the data are biased. This may lead to a decrease in the accuracy of the machine learning performance. Therefore, in this study, instead of calculating the projection from raw data to intermediate representation using only each user's raw data, each user randomly introduces data extracted from PubChem as projection data. As shown in Fig. 2, data collaboration analysis introduces the projection data (DCPd) and consists of the following four steps: 

Step 0: Each user randomly extracts anchor data ($X^{anc}$) and the projection data ($X^{p}_{i}$) from PubChem and shares only the former across all users. The anchor data is common, but the projection data is different across all users.

Step 1: Both the projection using each user's specific data ($f_{i}$) and that using their projection data ($f^{p}_{i}$) are calculated using dimensional reduction methods. When calculating $f_{i}$ and $f^{p}_{i}$, the sum of the dimension after transformation by $f_{i}$ and the dimension after transformation by $f^{p}_{i}$ must be smaller than the dimension of the user's specific raw data.
The features obtained by applying $f_{i}$ to each user's specific data and those acquired by applying $f^{p}_{i}$ to the data are then concatenated as an intermediate representation of each user's data. The intermediate representation of the anchor data is calculated similarly.

Step 2: The intermediate representation of each user's specific data is transformed into the collaboration representation.

Step 3: Machine learning using collaboration representation is performed similarly to conventional data collaboration analysis.

It is known that the performance of DC improves as the image spaces of the dimension reduction functions for each institution become closer \cite{imakura2021accuracy}. Generally, in non-IID cases, this relationship is disrupted, leading to an anticipated decrease in DC's performance. However, through the proposed method, it is expected that even in non-IID cases, bringing the image spaces of the dimension reduction functions for each institution closer together could contribute to the enhancement of DC's performance. The detailed algorithm for DCPd is shown in Algorithm 3.


\begin{figure}[h]
    \centering
    \includegraphics[keepaspectratio, scale=0.2]{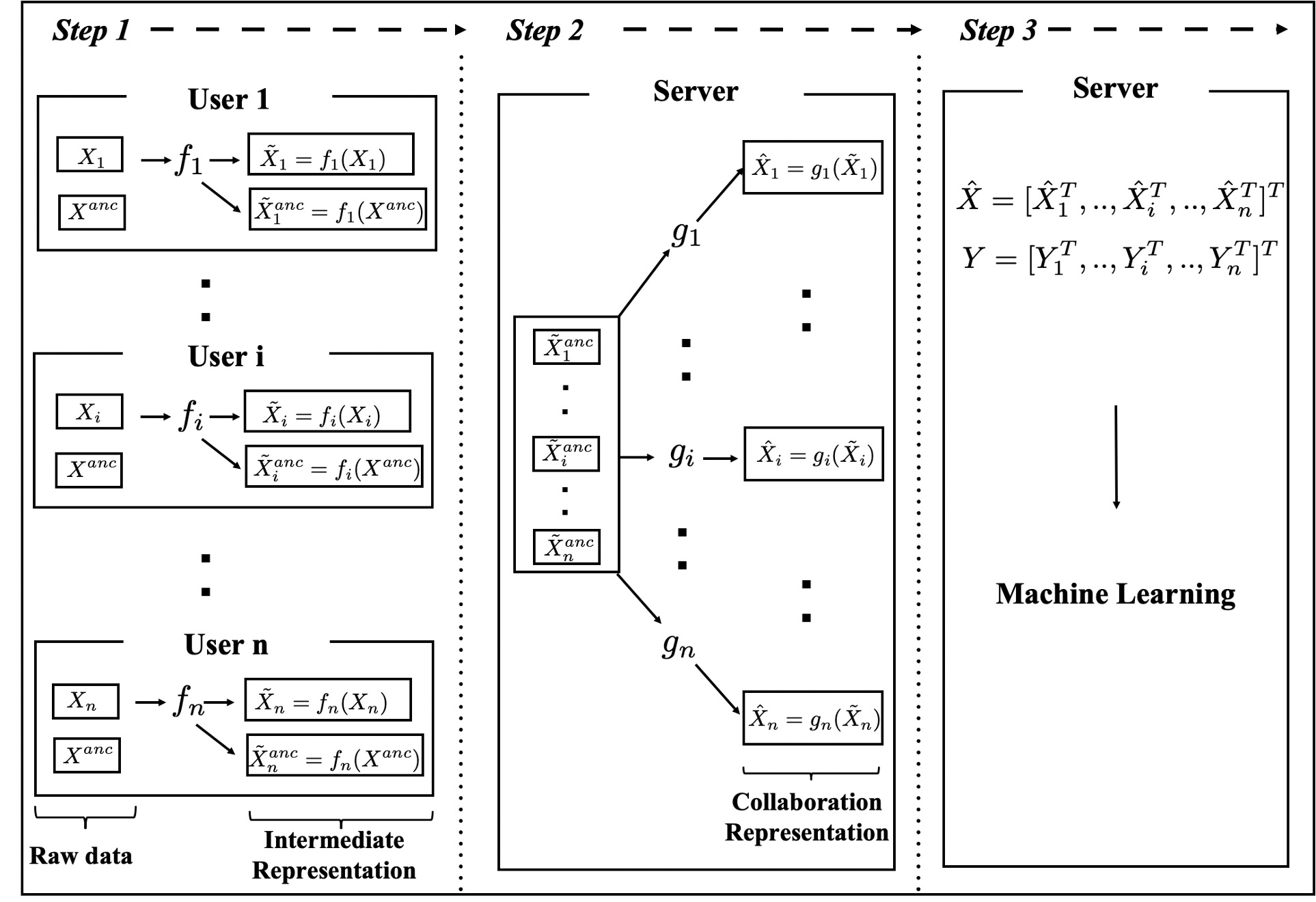}
  \caption{Overview of conventional data collaboration analysis (DC) ($X_{i}$ is user i's specific data, $X^{anc}$ is the common anchor data, $f_{i}$ is the projection of user i from raw data to intermediate representation, $\tilde{X}_{i}$ is the intermediate representation of user i's specific data, $\tilde{X}^{anc}_{i}$ is user i's intermediate representation of the anchor data, $g_{i}$ is the projection of user i from intermediate representation to collaboration representation, $\tilde{X}_{i}$ is a collaboration representation of user i's specific data, $\hat{X}^{anc}_{i}$ is user i's collaboration representation of the anchor data.)}\label{PreviousDC}
\end{figure}

 \begin{figure}[h]
    \centering
    \includegraphics[keepaspectratio, scale=0.4]{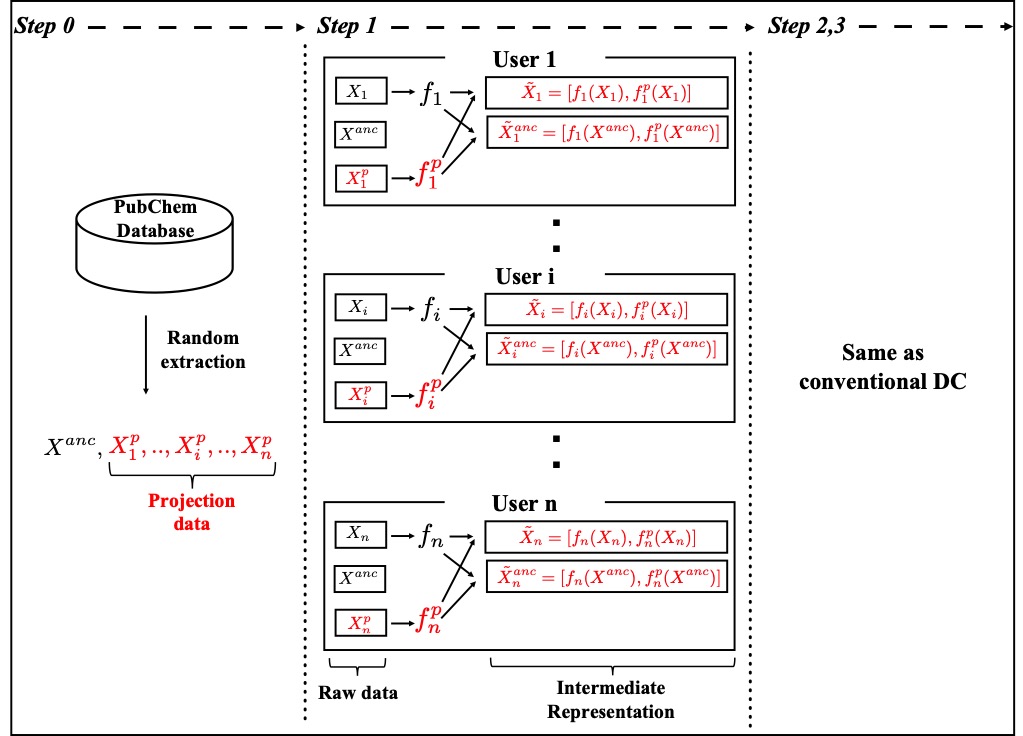}
  \caption{Overview of proposed data collaboration analysis with projection data (DCPd) where ($X^{anc}$ is the common anchor data, $X^{p}_{i}$ is user $i$'s projection data, $X_{i}$ is user $i$'s specific data, $f_{i}$ is the projection of user $i$'s specific data, $f^{p}_{i}$ is the projection of user $i$'s projection data, $\tilde{X}_{i}$ is the intermediate representation of user $i$'s specific data, $\tilde{X}^{anc}_{i}$ is user $i$'s intermediate representation of the anchor data. The red font indicates where they differ from the conventional DC.)}\label{DCPd}
\end{figure}

\renewcommand{\algorithmicrequire}{\textbf{Input:}}
\renewcommand{\algorithmicensure}{\textbf{Output:}}

\begin{algorithm}
\caption{Data collaboration analysis using projection data (DCPd), the proposed method}
\label{alg1}
\begin{algorithmic}[1]
\Require{$X_{i} \in R^{s_{i}\times{m}}, Y_{i} \in R^{s_{i}}$}
\Ensure{$Y^{test}=h([f_{i}(X^{test}),f_{i}^{p}(X^{test})]G_{i})$}
\State \textbf{USER SIDE:}
\State \{Step 0. Extraction of anchor data and projection data\}
\State {Extract $X^{anc} \in R^{a \times m}, X_{i}^{p} \in R^{b \times m}(i=1,2,...,n)$ from public database}
\State {Share $X^{anc}$ across all users}
\State \{Step 1. Construction of intermediate representation\}
\For{$i = 1,2,...,n$}
\State {Compute $f_{i},f_{i}^{p}$ based on dimensional reduction projection of $X^{anc},X_{i}^{p}$}
\State {Construct $\tilde{X}_{i}=[f_{i}(X_{i}),f_{i}^{p}(X_{i})], 
\tilde{X}_{i}^{anc}=[f_{i}(X^{anc}),f_{i}^{p}(X^{anc})]$}
\EndFor
\State {Centralize $\tilde{X}_{i}, \tilde{X}_{i}^{anc}$ and $Y_{i}$ for all $i$}
\State \textbf{SERVER SIDE:}
\State \{Step 2. Construction of collaboration representation\}
\State {Apply SVD to $\tilde{X}^{anc}=[\tilde{X}_{1}^{anc}, \tilde{X}_{2}^{anc},...,\tilde{X}_{n}^{anc}]$ 
$i.e.$  $\tilde{X}^{anc}=[U_{1},U_{2}]\begin{bmatrix}
\sum_{1} &  \\
 & \sum_{2} \\
\end{bmatrix}
\begin{bmatrix}
V_{1}^{T} \\
V_{2}^{T} \\
\end{bmatrix}$}
\For{$i = 1,2,..., n$}
\State {Compute $G_{i}=(\tilde{X}_{i}^{anc})^{\dag}U_{1}$}
\State {Compute $\hat{X}^{anc}=\tilde{X}_{i}G_{i}$}
\EndFor
\State {Set $\hat{X}=[\hat{X}_{1}^{T},\hat{X}_{2}^{T},...,\hat{X}_{n}^{T}]^{T}, Y=[Y_{1}^{T},Y_{2}^{T},...,Y_{n}^{T}]^{T}$}
\State \{Step 3. Machine learning with collaboration representation\}
\State {Construct model $h$ using a machine learning method with $\hat{X}$ as the training data and $Y$ as the ground truth $i.e.$ $Y \approx h(\hat{X})$}
\State {Obtain $Y^{test}=h([f_{i}(X^{test}), f_{i}^{p}(X^{test})]G_{i})$}
\end{algorithmic}
\end{algorithm}

\section{Experimental methods and results}
\label{sec:sample1}

\subsection{Datasets}
\label{sec:sample1}
To compare data collaboration analysis with federated learning, we used six datasets from Therapeutic Data Commons (TDC) \cite{huang2021therapeutics}: AMES \cite{xu2012silico}, CYP2D6\_Veith, CYP3A4\_Veith, CYP1A2\_Veith \cite{veith2009comprehensive}, HIV \cite{wu2018moleculenet}, Tox21\_SR-ARE \cite{mayr2016deeptox} (Table 1). In TDC, the default method for splitting training, validation, and the test data was set up. The datasets used in this study were converted into 2048-dimensional vectors using Morgan fingerprints (radius=2), followed by their splitting into training, validation, and test data.

\begin{table}[htbp]
\centering
\caption{Details of the public datasets used in this study.}
  \label{tab:Table 1}
  \begin{tabular}{ccccc}
    \hline
Dataset & Size & Task & Detail \\ 
    \hline
AMES & 7255 & Binary & Ames mutation assay \\ 
CYP2D6\_Veith & 13130 & Binary & Inhibition of CYP2D6 \\
CYP3A4\_Veith & 12328 & Binary & Inhibition of CYP3A4 \\
CYP1A2\_Veith & 12579 & Binary & Inhibition of CYP1A2 \\ 
HIV & 41127 & Binary & Inhibition of HIV replication \\
Tox21\_SR-ARE & 5932 & Binary & Nrf2/ARE signaling pathway assay \\
    \hline
\end{tabular}
\end{table}

\subsection{Method for generating anchor data}

\label{sec:sample1}
First, we compared the methods for creating anchor data using AMES dataset. At first, the training data was divided into four components in the IID setting to ensure that the number of samples was equal. In the centralized analysis, all data partitions were re-joined and used for training in a conventional manner (no anchor data was produced).
In the case of data collaboration analysis, the anchor data were (i) random values uniformly distributed in the range of 0–1 on 2048 dimensions, (ii) random values discretely distributed with 0, 1 on 2048 dimensions, and (iii) randomly selected among those with PubChem CID of 1–12000 and converted into 2048-dimensional vectors using Morgan fingerprints. Dimensional reduction for the creation of intermediate representation was conducted using truncated singular value decomposition (SVD). The dimension of the intermediate representation was 500, and the dimension of the collaborative representation was 100. The neural networks, as shown in Table 2, were used for training using the training data. During each epoch, the validation loss was checked using validation data, with a patience value set to 10, determining the stopping point for the training process. ROC-AUC and PR-AUC were evaluated on the test data using this model. The experiment was repeated 5 times with different partitioning of the training data, and the average values of ROC-AUC and PR-AUC were calculated. Note that the experiment was conducted only once for the centralized analysis because the results did not change, even if the division of the training data was changed.

As described in Table 3, the investigation of the anchor data creation methods revealed that the anchor data extracted from PubChem (DC\_acPC\_2000) resulted in superior performance for both ROC-AUC and PR-AUC compared to the anchor data created using a discrete distribution of 0 or 1 (DC\_ac0or1\_2000) or a uniform distribution of 0 to 1 (DC\_ac0-1\_2000).

\begin{table}[htbp]
\centering
\caption{Parameters of neural network used in this study.}
  \label{tab:Table 2}
  \scalebox{1}{
  \begin{tabular}{cccccccc}
  \hline
Parameters & Values  \\ 
  \hline
Number of hidden layers & 2  \\ 
Number of neurons in each layer & 2000, 1000  \\ 
Dropout (\%) in each Layer & 40, 40  \\ 
Minibatch size & 25  \\
Optimizer & Adam \cite{kingma2014adam}  \\
Learning rate & 0.00002  \\
Early stopping & Val\_loss (patience=10)  \\
Max epochs & 300  \\
    \hline
\end{tabular}}
\end{table}

\begin{table}[htbp]
\centering
\caption{Mean ROC-AUC and PR-AUC for change of the creation method of the anchor data for the AMES dataset in IID settings (Mean ± standard error).}
  \label{tab:Table 1}
  \begin{tabular}{cccccc}
    \hline
Method & ROC-AUC & PR-AUC \\ 
    \hline
Centralized & 0.885 & 0.903 \\ 
DC\_ac0-1\_2000 & 0.814±0.005 & 0.841±0.006 \\ 
DC\_ac0or1\_2000 & 0.790±0.006 & 0.820±0.005 \\ 
DC\_acPC\_2000 & 0.844±0.003 & 0.865±0.004 \\ 
    \hline
\end{tabular}
\end{table}

\subsection{Introduction of projection data}
\label{sec:sample1}
Next, we introduced the projection data to the data collaboration analysis. At first, compound data with PubChem CIDs between 20001 and 120000 were extracted and converted to 2048-dimensional Morgan fingerprints. The projection data were randomly selected for each user from these data. Each user then calculated the projection $f_{i}$ for 100 dimensions from the raw data and the projection $f^{p}_{i}$ for 100 dimensions from the projection data using truncated SVD. The number of projection data was varied from 2000 (DCPd\_2000), 5000 (DCPd\_5000), 10000 (DCPd\_10000), 20000 (DCPd\_20000), and 40000 (DCPd\_40000) to compare the centralized analysis with the data collaboration analysis. In previous experiments, the number of samples for each user and the ratio of the labels were set at random (IID settings). However, in this experiment, the training data was split under extreme non-IID settings, where two of the four users equally shared label-0 samples, and label-1 samples were equally split among the remaining two users. The experiment was conducted based on the approach described in Section 3.2, except that 3000 anchor data from PubChem were used. The dimension of the intermediate representation was 200 (100 dimensions were calculated using $f_{i}$ and the other 100 dimensions were calculated using $f^{p}_{i}$ in the case of DCPd), the dimension of the collaboration representation was fixed at 100, and the method of creating intermediate representations based on projection data was utilized.

The results showed that as the number of projection data was increased, the classification accuracy increased up to 20000 projection data, after which the change was negligible (Table 4).

\begin{table}[htbp]
\centering
\caption{Mean ROC-AUC and PR-AUC when the number of projection data was changed for the AMES dataset in non-IID settings (Mean ± standard error).}
  \label{tab:Table 2}
  \begin{tabular}{cccccc}
    \hline
Method & ROC-AUC & PR-AUC \\ 
    \hline
Centralized & 0.885 & 0.903 \\ 
DC & 0.800±0.006 & 0.818±0.006 \\ 
DCPd\_2000 & 0.822±0.006 & 0.834±0.007 \\ 
DCPd\_5000 & 0.836±0.002 & 0.849±0.004 \\ 
DCPd\_10000 & 0.836±0.001 & 0.852±0.002 \\ 
DCPd\_20000 & 0.840±0.002 & 0.856±0.004 \\ 
DCPd\_40000 & 0.842±0.002 & 0.857±0.003 \\ 
    \hline
\end{tabular}
\end{table}

\subsection{Comparison of FedAvg, DC, and DCPd with IID compound datasets}
\label{sec:sample1}
Next, comparisons of centralized analysis, federated learning, normal data collaboration analysis (DC), and data collaboration analysis based on projection data (DCPd) were performed using the IID partitioning method on all the datasets except AMES. The training data were split using the IID method, similar to the experiments in Section 3.2. For the federated learning method, we used the FedAvg algorithm described in Section 2.1, whose parameters are described in Table 5. The neural network outlined in Table 2 served as the machine learning method for federated learning.

For centralized analysis, DC and DCPd, the same method was used as in the experiment in Section 3.3, and the number of projection data for DCPd was set to 20,000.

As described in Table 6 and 7, DC and DCPd exhibited higher ROC-AUC and PR-AUC compared to FedAvg in CYP2D6, the HIV, and Tox21\_SR-ARE datasets. FedAvg was superior to DC and DCPd in the CYP3A4 and CYP1A2 datasets. Overall, there was no significant difference in their accuracy.

\begin{table}[htbp]
\centering
\caption{Parameters of federated averaging used in this study.}
  \label{tab:Table 5}
  \scalebox{1}{
  \begin{tabular}{cccccccc}
  \hline
Parameters & Values  \\ 
  \hline
Number of epoch per each round & 1  \\
Early stopping & Val\_loss (patience=10)  \\
Max rounds & 300  \\
    \hline
\end{tabular}}
\end{table}

\begin{table}[htbp]
\centering
\begin{threeparttable}
\caption{Mean ROC-AUC of centralized analysis, FedAvg, DC and DCPd of five compound datasets (CYP2D6, CYP3A4, CYP1A2, HIV, Tox21\_SR-ARE) in IID settings (Mean ± standard error).}
  \label{tab:Table 5}
  
  \begin{tabular}{cccccc}
    \hline
Method & CYP2D6 & CYP3A4 & CYP1A2 & HIV & SR-ARE \\ 
    \hline
Centralized & 0.844 & 0.876 & 0.908 & 0.769 & 0.745\\ 
FedAvg & \makecell{0.789 \\ ±0.002} & \makecell{\textbf{0.837} \\ \textbf{±0.002}} & \makecell{\textbf{0.883} \\ \textbf{±0.001}} & \makecell{0.666 \\ ±0.004} & \makecell{0.654 \\ ±0.008} \\ 
DC & \makecell{\textbf{0.805} \\ \textbf{±0.001}} & \makecell{0.826 \\ ±0.002} & \makecell{0.869 \\ ±0.001} & \makecell{\textbf{0.735} \\ \textbf{±0.005}} & \makecell{0.674 \\ ±0.005} \\ 
DCPd & \makecell{0.799 \\ ±0.001} & \makecell{0.825 \\ ±0.001} & \makecell{0.870 \\ ±0.002} & \makecell{0.729 \\ ±0.002} & \makecell{\textbf{0.676} \\ \textbf{±0.004}} \\ 
    \hline
    \end{tabular}
\begin{tablenotes}
      \small
      \item \textit{Note:} Bold values denote the best results among FedAvg, DC, and DCPd.
\end{tablenotes}
\end{threeparttable}
\end{table}

\begin{table}[htbp]
\centering
\begin{threeparttable}
\caption{Mean PR-AUC of centralized analysis, FedAvg, DC and DCPd for five compound datasets (CYP2D6, CYP3A4, CYP1A2, HIV, Tox21\_SR-ARE) in IID settings (Mean ± standard error).}
  \label{tab:Table 6}
  \begin{tabular}{cccccc}
    \hline
Method & CYP2D6 & CYP3A4 & CYP1A2 & HIV & SR-ARE \\ 
    \hline
Centralized & 0.648 & 0.836 & 0.900 & 0.400 & 0.373\\ 
FedAvg & \makecell{0.854 \\ ±0.002} & \makecell{\textbf{0.795} \\ \textbf{±0.002}} & \makecell{\textbf{0.873} \\ \textbf{±0.001}} & \makecell{0.225 \\ ±0.005} & \makecell{0.293 \\ ±0.007} \\ 
DC & \makecell{\textbf{0.592} \\ \textbf{±0.001}} & \makecell{0.778 \\ ±0.002} & \makecell{0.863 \\ ±0.002} & \makecell{0.310 \\ ±0.007} & \makecell{0.283 \\ ±0.008} \\ 
DCPd & \makecell{0.591 \\ ±0.003} & \makecell{0.774 \\ ±0.001} & \makecell{0.861 \\ ±0.002} & \makecell{\textbf{0.311} \\ \textbf{±0.004}} & \makecell{0.285 \\ ±0.007} \\ 
    \hline
    \end{tabular}
\begin{tablenotes}
      \small
      \item \textit{Note:} Bold values denote the best results among FedAvg, DC, and DCPd.
\end{tablenotes}
\end{threeparttable}
\end{table}

\subsection{Comparison of FedAvg, DC, and DCPd with Non-IID compound datasets}
\label{sec:sample1}
Next, comparisons of centralized analysis, FedAvg, DC, and DCPd were conducted using the non-IID partitioning method for all datasets except AMES. In the non-IID settings with four users, two of which had label 0 and two of which had only label 1 as in Section 3.3. The same methods were used as in Section 3.4, except for the splitting of the datasets.

The results showed that the ROC-AUC and PR-AUC of DCPd were best and those of FedAvg were worst for all the datasets (Table 8 and 9).

\begin{table}[htbp]
\centering
\begin{threeparttable}
\caption{Mean ROC-AUC of centralized analysis, FedAvg, DC, and DCPd for five compound datasets (CYP2D6, CYP3A4, CYP1A2, HIV, Tox21\_SR-ARE) of non-IID settings (Mean ± standard error).}
\label{tab:Table 7}
\begin{tabular}{cccccc}
\hline
Method & CYP2D6 & CYP3A4 & CYP1A2 & HIV & SR-ARE \\
\hline
Centralized & 0.844 & 0.876 & 0.908 & 0.769 & 0.745\\
FedAvg & \makecell{0.489 \\ ±0.007} & \makecell{0.343 \\ ±0.008} & \makecell{0.705 \\ ±0.068} & \makecell{0.473 \\ ±0.002} & \makecell{0.444 \\ ±0.004} \\
DC & \makecell{0.755 \\ ±0.007} & \makecell{0.781 \\ ±0.006} & \makecell{0.833 \\ ±0.002} & \makecell{0.707 \\ ±0.004} & \makecell{0.653 \\ ±0.004} \\
DCPd & \makecell{\textbf{0.780} \\ \textbf{±0.006}} & \makecell{\textbf{0.818} \\ \textbf{±0.001}} & \makecell{\textbf{0.862} \\ \textbf{±0.002}} & \makecell{\textbf{0.745} \\ \textbf{±0.003}} & \makecell{\textbf{0.675} \\ \textbf{±0.007}} \\
\hline
\end{tabular}
\begin{tablenotes}
\small
\item \textit{Note:} Bold values denote the best results among FedAvg, DC, and, DCPd.
\end{tablenotes}
\end{threeparttable}
\end{table}

\begin{table}[htbp]
\centering
\begin{threeparttable}
\caption{Mean PR-AUC of centralized analysis, FedAvg, DC, and DCPd for five compound datasets (CYP2D6, CYP3A4, CYP1A2, HIV, Tox21\_SR-ARE) of non-IID settings (Mean ± standard error).}
\label{tab:Table 8}
\begin{tabular}{cccccc}
\hline
Method & CYP2D6 & CYP3A4 & CYP1A2 & HIV & SR-ARE \\
\hline
Centralized & 0.648 & 0.836 & 0.900 & 0.400 & 0.373\\
FedAvg & \makecell{0.171 \\ ±0.002} & \makecell{0.314 \\ ±0.003} & \makecell{0.667 \\ ±0.080} & \makecell{0.033 \\ ±0.000} & \makecell{0.131 \\ ±0.002} \\
DC & \makecell{0.501 \\ ±0.007} & \makecell{0.724 \\ ±0.005} & \makecell{0.815 \\ ±0.002} & \makecell{0.256 \\ ±0.006} & \makecell{0.258 \\ ±0.010} \\
DCPd & \makecell{\textbf{0.544} \\ \textbf{±0.009}} & \makecell{\textbf{0.753} \\ \textbf{±0.002}} & \makecell{\textbf{0.849} \\ \textbf{±0.002}} & \makecell{\textbf{0.310} \\ \textbf{±0.008}} & \makecell{\textbf{0.288} \\ \textbf{±0.010}} \\
\hline
\end{tabular}
\begin{tablenotes}
\small
\item \textit{Note:} Bold values denote the best results among FedAvg, DC, and DCPd.
\end{tablenotes}
\end{threeparttable}
\end{table}

\subsection{Comparison of FedAvg, DC, and DCPd with compound datasets for varying label proportions}
\label{sec:sample1}
We introduced an index of label bias, $r$, and assigned the training data with label 0 to User 1--4 in the proportions of $(25+25r)\%$, $(25+25r)\%$, $(25-25r)\%$, $(25-25r)\%$ and that with label 1 to User 1--4 in the proportions of $(25-25r)\%$, $(25-25r)\%$, $(25+25r)\%$,  $(25+25r)\%$, as shown in Table 10. Thus, there was no label bias when $r$ was 0 and the label bias was greatest when $r$ was 1. We compared federated learning, DC, and DCPd on CYP2D6, CYP3A4, CYP1A2, Tox21\_SR-ARE datasets by varying $r$ as 0, 0.2, 0.4, 0.6, 0.8, 0.85, 0.9, 0.95, and 1.0.

As shown in Fig. 3, the classification accuracies of DCPd did not decrease significantly with the increase of the label bias of each user's data for all five datasets used in this experiment, although those of DC decreased slightly, and those of FedAvg decreased significantly.

\begin{table}[htbp]
\caption{Splitting of the training data for each user with $r$}
\centering
  \label{tab:Table 9}
  \begin{tabular}{cccccc}
    \hline
Data & User 1, 2 & User 3, 4 & Total \\ 
    \hline
\makecell{Training data \\ (Label:0)} & $(25+25r)\%$ & $(25-25r)\%$ & $100\%$ \\ 
\makecell{Training data \\ (Label:1)} & $(25-25r)\%$ & $(25+25r)\%$ & $100\%$ \\
    \hline
    \end{tabular}
\end{table}

\begin{figure}[htbp]

  \begin{minipage}[b]{0.322\linewidth}
    \centering
    \includegraphics[keepaspectratio, scale=0.23]{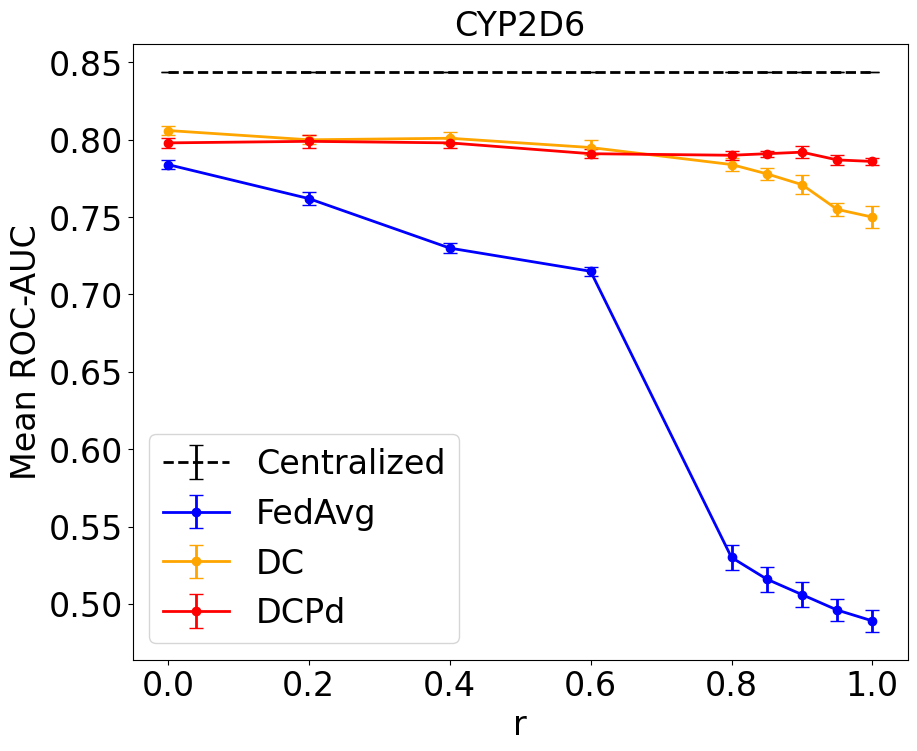}
    \subcaption{\scriptsize ROC-AUC for CYP2D6}\label{CYP2D6_ROC-AUC}
  \end{minipage}
  \begin{minipage}[b]{0.322\linewidth}
    \centering
    \includegraphics[keepaspectratio, scale=0.23]{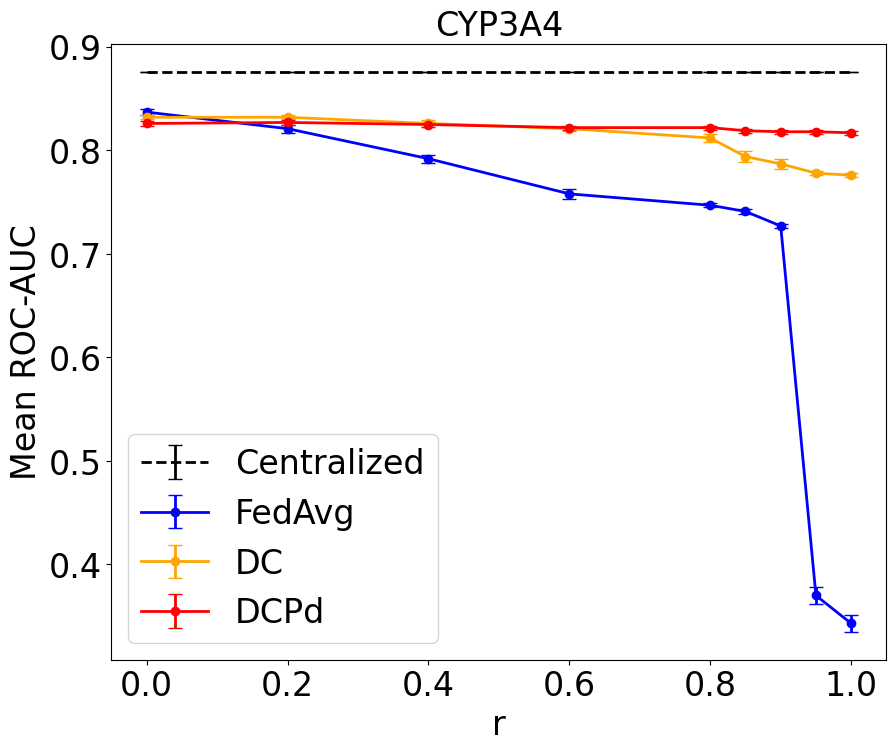}
    \subcaption{\scriptsize ROC-AUC for CYP3A4}\label{CYP3A4_ROC-AUC}
  \end{minipage}
  \begin{minipage}[b]{0.322\linewidth}
    \centering
    \includegraphics[keepaspectratio, scale=0.23]{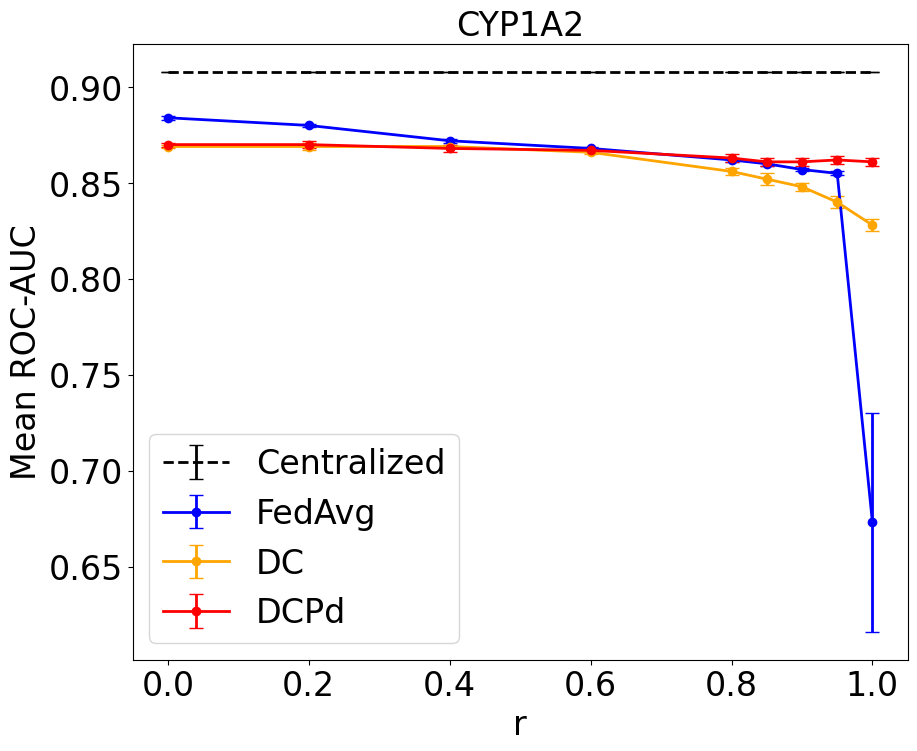}
    \subcaption{\scriptsize ROC-AUC for CYP1A2}\label{CYP1A2_ROC-AUC}
  \end{minipage}
  
  \begin{minipage}[b]{0.322\linewidth}
    \centering
    \includegraphics[keepaspectratio, scale=0.23]{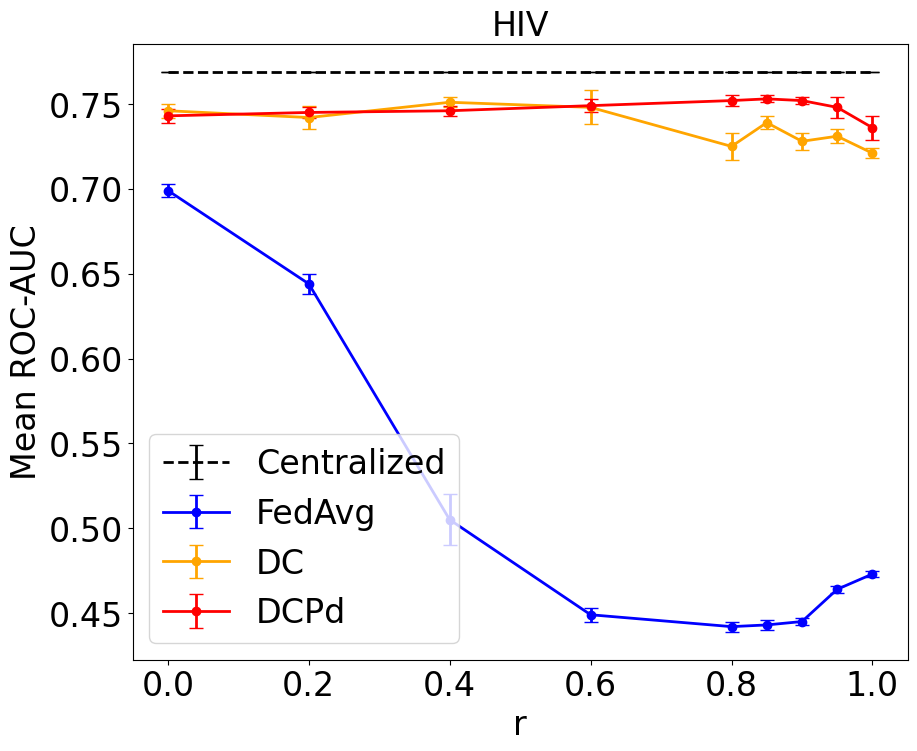}
    \subcaption{\scriptsize ROC-AUC for HIV}\label{HIV_ROC-AUC}
  \end{minipage}
  \begin{minipage}[b]{0.322\linewidth}
    \centering
    \includegraphics[keepaspectratio, scale=0.23]{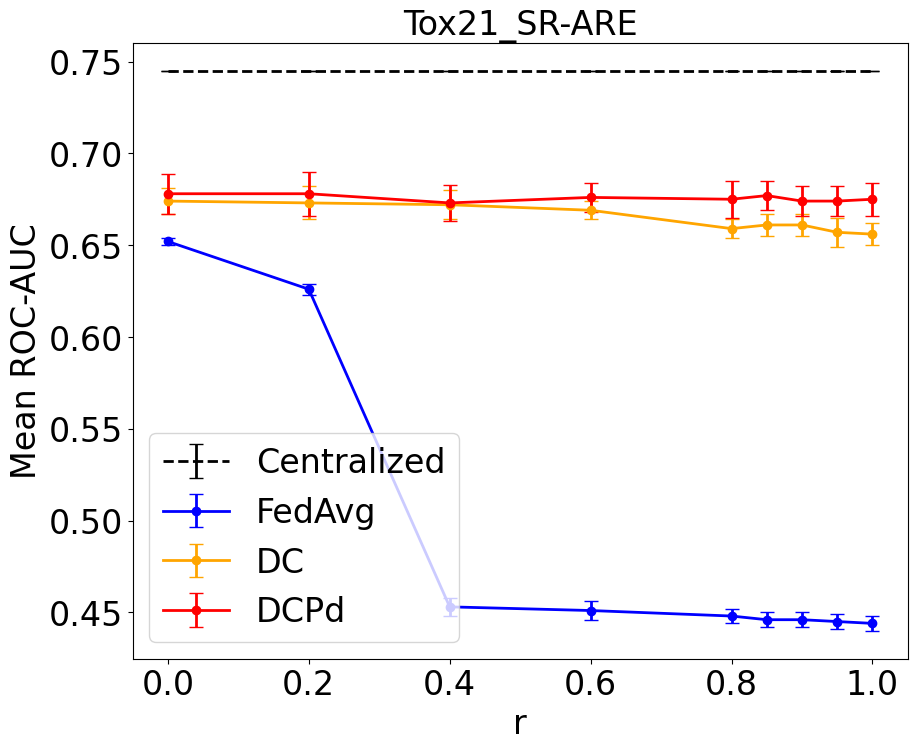}
    \subcaption{\scriptsize ROC-AUC for SR-ARE}\label{SR-ARE_ROC-AUC}
  \end{minipage}
  
  \begin{minipage}[b]{0.322\linewidth}
    \centering
    \includegraphics[keepaspectratio, scale=0.23]{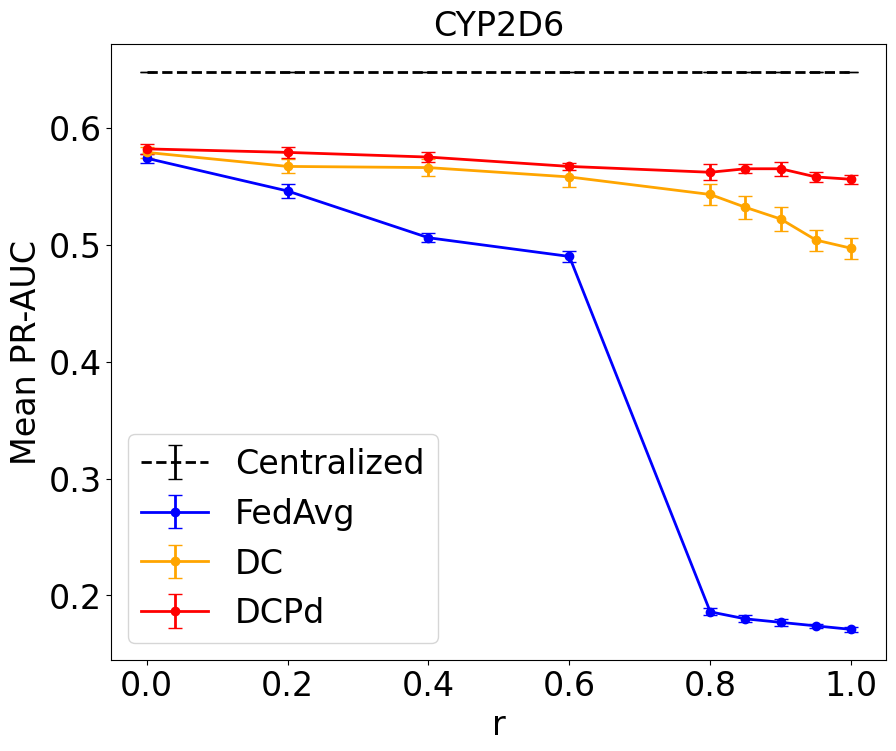}
    \subcaption{\scriptsize PR-AUC for CYP2D6}\label{CYP2D6_PR-AUC}
  \end{minipage}
  \begin{minipage}[b]{0.322\linewidth}
    \centering
    \includegraphics[keepaspectratio, scale=0.23]{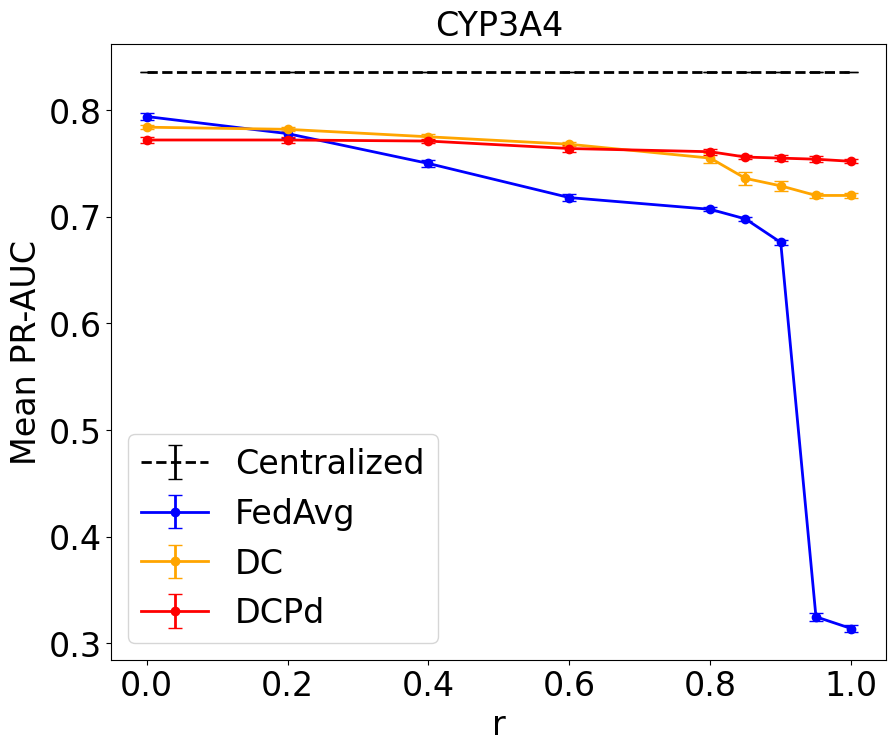}
    \subcaption{\scriptsize PR-AUC for CYP3A4}\label{CYP3A4_PR-AUC}
  \end{minipage}
  \begin{minipage}[b]{0.322\linewidth}
    \centering
    \includegraphics[keepaspectratio, scale=0.23]{CYP2D6_PR-AUC.jpg}
    \subcaption{\scriptsize PR-AUC for CYP1A2}\label{CYP1A2_PR-AUC}
  \end{minipage}

  \begin{minipage}[b]{0.322\linewidth}
    \centering
    \includegraphics[keepaspectratio, scale=0.23]{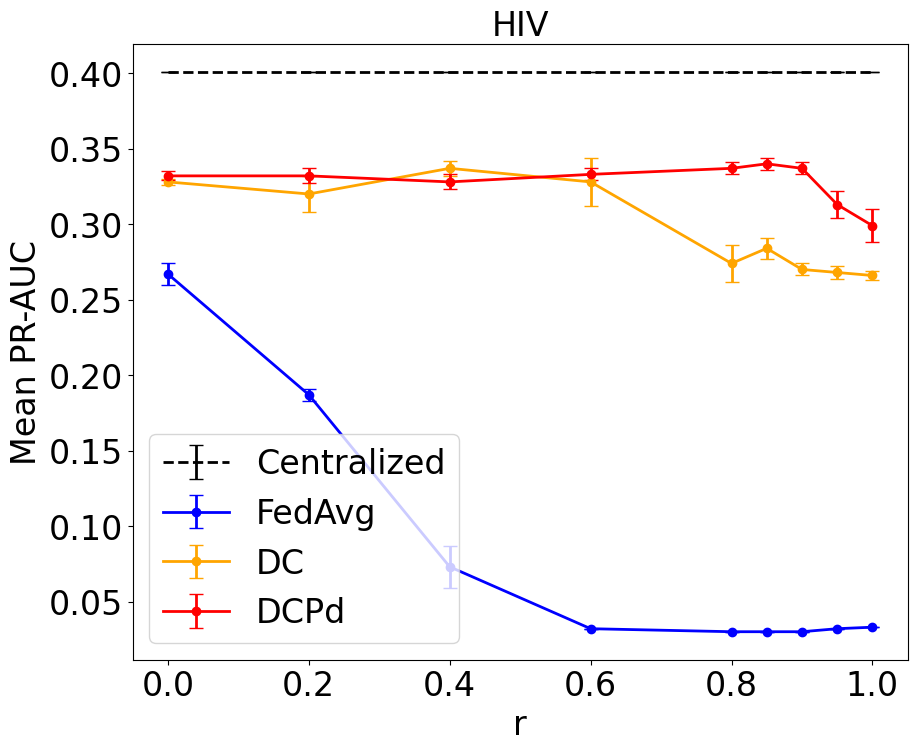}
    \subcaption{\scriptsize PR-AUC for HIV}\label{HIV_PR-AUC}
  \end{minipage}
  \begin{minipage}[b]{0.322\linewidth}
    \centering
    \includegraphics[keepaspectratio, scale=0.23]{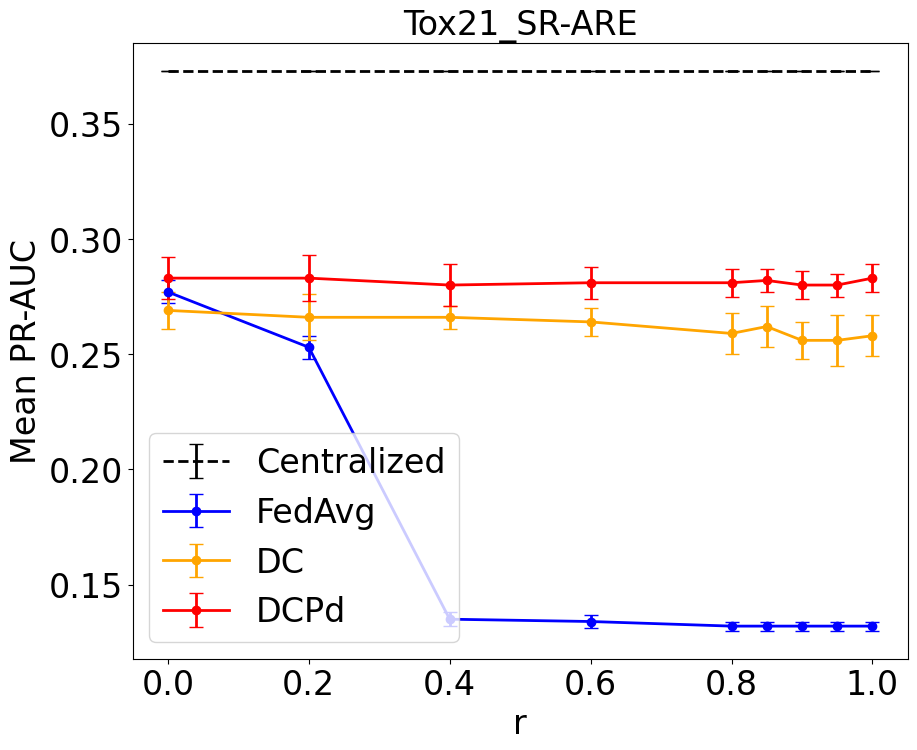}
    \subcaption{\scriptsize PR-AUC for SR-ARE}\label{SR-ARE_PR-AUC}
  \end{minipage}

  \caption{Classification performance of centralized analysis (Centralized), Federated Averaging (FedAvg), data collaboration analysis (DC), and the proposed method (DCPd) for five datasets when the index of label bias $r$ of data partition is varied from 0 to 1 (Mean ± standard error). }\label{ラベル}
\end{figure}

\section{Discussion}
\label{sec:sample1}
In this study, we demonstrated the applicability of data collaboration analysis to compound data by comparing federated learning and data collaboration analysis for IID and non-IID settings. The objective was to improve the accuracy of machine learning in non-IID settings since poor performance is one of the key challenges in federated learning. The proposed method is expected to be applicable to privacy-preserving machine learning for various non-IID data scenarios, including collaborative drug discovery initiatives among multiple pharmaceutical companies with biased compound libraries, where the implementation of federated learning has previously been challenging.

The results revealed that a higher classification accuracy was achieved when public data were used as the anchor data compared to existing anchor data creation methods, such as using a uniform distribution, where the minimum and maximum values are used as the lower and upper limits. A possible explanation is that using data with a distribution similar to that of the actual compound data as the anchor data contributes to the creation of a more appropriate representation in data collaboration analysis. It was also determined that the performance of FedAvg, the main method of federated learning, and data collaboration analysis are similar for IID settings. However, the performance for data collaboration analysis was superior compared to FedAvg for non-IID settings. This is because the training of FedAvg was performed using only data with one label, which makes it difficult to appropriately update the parameters of the neural network. In contrast, data with both 0 and 1 labels are learned simultaneously in data collaboration analysis, which makes it easier to update the parameters of the neural network. The third finding of this study is that DCPd, which uses the projection of dimensional reduction of public data to create intermediate representations for data collaboration analysis, greatly improves classification accuracy for non-IID settings compared to normal DC. This is because an appropriate projection for creating an intermediate representation cannot be obtained from each user's data only, owing to data bias in non-IID settings. However, the projection is calculated by combining the projection created from each user’s data and the unlabelled public data.

The first limitation of the proposed method is that it cannot be applied to compound graph data because data collaboration analysis has not been applied to graph structures thus far. However, federated learning has been used for the graph data of compounds. To address this problem, it is necessary to develop a method for applying data collaboration analysis to graph structures. The second limitation is that it can only be applied to fields where unlabelled public data is currently available. To solve this problem, this method should be combined with the creation of artificial data using GAN or other methods.

\section{Conclusion}
\label{sec:sample1}
In this study, we applied data collaboration analysis to compound datasets and compared it with FedAvg in IID or non-IID settings. The results showed that DC has a higher classification accuracy than FedAvg in non-IID settings. In addition, we proposed a method that used projections created from unlabelled public data to create intermediate representations in data collaboration analysis (DCPd). It was shown that DCPd facilitates higher classification accuracy in non-IID settings. Non-IID settings are common in a wide variety of fields. Therefore, the results of this study can be applied to privacy-preserving distributed learning in non-IID settings.

There are three major novel aspects of this study: 1) it is the first application of data collaboration analysis to chemical compound data; 2) data collaboration analysis was compared to federated learning for non-IID setting; 3) we showed that using public data to create intermediate representations for data collaboration analysis increased the performance of machine learning for non-IID settings. It was also shown that the difference in the classification accuracy between the IID settings and the non-IID settings is extremely small. In federated learning, the improvement of the performance of machine learning for non-IID settings has been investigated, but at present, the difference in the performance between the IID settings and the non-IID is large.

In future research, we will consider extending the proposed method for distributed chemical data analysis to other data structures, such as compound graph data or text data, as well as a variety of downstream tasks including clustering, generation, or anomaly detection. The proposed method is currently only applicable to the cases for which public databases such as PubChem are available. Therefore, the generation of projection data via augmentation or other methods for situations where public databases are not available remains a challenge.


\printbibliography

@article{wouters2020estimated,
  title={Estimated research and development investment needed to bring a new medicine to market, 2009-2018},
  author={Wouters, Olivier J and McKee, Martin and Luyten, Jeroen},
  journal={Jama},
  volume={323},
  number={9},
  pages={844--853},
  year={2020},
  publisher={American Medical Association}
}

@article{hughes2011principles,
  title={Principles of early drug discovery},
  author={Hughes, James P and Rees, Stephen and Kalindjian, S Barrett and Philpott, Karen L},
  journal={British journal of pharmacology},
  volume={162},
  number={6},
  pages={1239--1249},
  year={2011},
  publisher={Wiley Online Library}
}

@article{mayr2016deeptox,
  title={DeepTox: toxicity prediction using deep learning},
  author={Mayr, Andreas and Klambauer, G{\"u}nter and Unterthiner, Thomas and Hochreiter, Sepp},
  journal={Frontiers in Environmental Science},
  volume={3},
  pages={80},
  year={2016},
  publisher={Frontiers Media SA}
}

@article{burki2019pharma,
  title={Pharma blockchains AI for drug development},
  author={Burki, Talha},
  journal={The Lancet},
  volume={393},
  number={10189},
  pages={2382},
  year={2019},
  publisher={Elsevier}
}

@article{zhao2018federated,
  title={Federated learning with non-iid data},
  author={Zhao, Yue and Li, Meng and Lai, Liangzhen and Suda, Naveen and Civin, Damon and Chandra, Vikas},
  journal={arXiv preprint arXiv:1806.00582},
  year={2018}
}

@article{zhu2021federated,
  title={Federated learning on non-IID data: A survey},
  author={Zhu, Hangyu and Xu, Jinjin and Liu, Shiqing and Jin, Yaochu},
  journal={Neurocomputing},
  volume={465},
  pages={371--390},
  year={2021},
  publisher={Elsevier}
}

@article{tanner1987calculation,
  title={The calculation of posterior distributions by data augmentation},
  author={Tanner, Martin A and Wong, Wing Hung},
  journal={Journal of the American statistical Association},
  volume={82},
  number={398},
  pages={528--540},
  year={1987},
  publisher={Taylor \& Francis}
}

@article{hinton2015distilling,
  title={Distilling the knowledge in a neural network},
  author={Hinton, Geoffrey and Vinyals, Oriol and Dean, Jeff and others},
  journal={arXiv preprint arXiv:1503.02531},
  volume={2},
  number={7},
  year={2015}
}

@article{yoon2021fedmix,
  title={Fedmix: Approximation of mixup under mean augmented federated learning},
  author={Yoon, Tehrim and Shin, Sumin and Hwang, Sung Ju and Yang, Eunho},
  journal={arXiv preprint arXiv:2107.00233},
  year={2021}
}

@article{li2022federated,
  title={Federated Learning with GAN-based Data Synthesis for Non-IID Clients},
  author={Li, Zijian and Shao, Jiawei and Mao, Yuyi and Wang, Jessie Hui and Zhang, Jun},
  journal={arXiv preprint arXiv:2206.05507},
  year={2022}
}

@article{li2019fedmd,
  title={Fedmd: Heterogenous federated learning via model distillation},
  author={Li, Daliang and Wang, Junpu},
  journal={arXiv preprint arXiv:1910.03581},
  year={2019}
}

@article{lin2020ensemble,
  title={Ensemble distillation for robust model fusion in federated learning},
  author={Lin, Tao and Kong, Lingjing and Stich, Sebastian U and Jaggi, Martin},
  journal={Advances in Neural Information Processing Systems},
  volume={33},
  pages={2351--2363},
  year={2020}
}

@article{wang2009pubchem,
  title={PubChem: a public information system for analyzing bioactivities of small molecules},
  author={Wang, Yanli and Xiao, Jewen and Suzek, Tugba O and Zhang, Jian and Wang, Jiyao and Bryant, Stephen H},
  journal={Nucleic acids research},
  volume={37},
  number={suppl\_2},
  pages={W623--W633},
  year={2009},
  publisher={Oxford University Press}
}

@misc{pence2010chemspider,
  title={ChemSpider: an online chemical information resource},
  author={Pence, Harry E and Williams, Antony},
  year={2010},
  publisher={ACS Publications}
}

@article{gaulton2012chembl,
  title={ChEMBL: a large-scale bioactivity database for drug discovery},
  author={Gaulton, Anna and Bellis, Louisa J and Bento, A Patricia and Chambers, Jon and Davies, Mark and Hersey, Anne and Light, Yvonne and McGlinchey, Shaun and Michalovich, David and Al-Lazikani, Bissan and others},
  journal={Nucleic acids research},
  volume={40},
  number={D1},
  pages={D1100--D1107},
  year={2012},
  publisher={Oxford University Press}
}

@article{huang2021therapeutics,
  title={Therapeutics data commons: Machine learning datasets and tasks for drug discovery and development},
  author={Huang, Kexin and Fu, Tianfan and Gao, Wenhao and Zhao, Yue and Roohani, Yusuf and Leskovec, Jure and Coley, Connor W and Xiao, Cao and Sun, Jimeng and Zitnik, Marinka},
  journal={arXiv preprint arXiv:2102.09548},
  year={2021}
}

@article{xu2012silico,
  title={In silico prediction of chemical Ames mutagenicity},
  author={Xu, Congying and Cheng, Feixiong and Chen, Lei and Du, Zheng and Li, Weihua and Liu, Guixia and Lee, Philip W and Tang, Yun},
  journal={Journal of chemical information and modeling},
  volume={52},
  number={11},
  pages={2840--2847},
  year={2012},
  publisher={ACS Publications}
}

@article{veith2009comprehensive,
  title={Comprehensive characterization of cytochrome P450 isozyme selectivity across chemical libraries},
  author={Veith, Henrike and Southall, Noel and Huang, Ruili and James, Tim and Fayne, Darren and Artemenko, Natalia and Shen, Min and Inglese, James and Austin, Christopher P and Lloyd, David G and others},
  journal={Nature biotechnology},
  volume={27},
  number={11},
  pages={1050--1055},
  year={2009},
  publisher={Nature Publishing Group}
}

@article{wu2018moleculenet,
  title={MoleculeNet: a benchmark for molecular machine learning},
  author={Wu, Zhenqin and Ramsundar, Bharath and Feinberg, Evan N and Gomes, Joseph and Geniesse, Caleb and Pappu, Aneesh S and Leswing, Karl and Pande, Vijay},
  journal={Chemical science},
  volume={9},
  number={2},
  pages={513--530},
  year={2018},
  publisher={Royal Society of Chemistry}
}

@article{imakura2020data,
  title={Data Collaboration Analysis Framework Using Centralization of Individual Intermediate Representations for Distributed Data Sets},
  author={Imakura, Akira and Sakurai, Tetsuya},
  journal={ASCE-ASME Journal of Risk and Uncertainty in Engineering Systems, Part A: Civil Engineering},
  volume={6},
  number={2},
  pages={04020018},
  year={2020},
  publisher={American Society of Civil Engineers}
}

@inproceedings{imakura2021collaborative,
  title={Collaborative data analysis: non-model sharing-type machine learning for distributed data},
  author={Imakura, Akira and Ye, Xiucai and Sakurai, Tetsuya},
  booktitle={Pacific Rim Knowledge Acquisition Workshop},
  pages={14--29},
  year={2021},
  organization={Springer}
}

@article{bogdanova2020federated,
  title={Federated learning system without model sharing through integration of dimensional reduced data representations},
  author={Bogdanova, Anna and Nakai, Akie and Okada, Yukihiko and Imakura, Akira and Sakurai, Tetsuya},
  journal={arXiv preprint arXiv:2011.06803},
  year={2020}
}

@article{imakura2021interpretable,
  title={Interpretable collaborative data analysis on distributed data},
  author={Imakura, Akira and Inaba, Hiroaki and Okada, Yukihiko and Sakurai, Tetsuya},
  journal={Expert Systems with Applications},
  volume={177},
  pages={114891},
  year={2021},
  publisher={Elsevier}
}

@article{hung2021qsar,
  title={QSAR modeling without descriptors using graph convolutional neural networks: the case of mutagenicity prediction},
  author={Hung, Chiakang and Gini, Giuseppina},
  journal={Molecular diversity},
  volume={25},
  number={3},
  pages={1283--1299},
  year={2021},
  publisher={Springer}
}

@article{karpov2020transformer,
  title={Transformer-CNN: Swiss knife for QSAR modeling and interpretation},
  author={Karpov, Pavel and Godin, Guillaume and Tetko, Igor V},
  journal={Journal of Cheminformatics},
  volume={12},
  number={1},
  pages={1--12},
  year={2020},
  publisher={BioMed Central}
}

@article{myint2012molecular,
  title={Molecular fingerprint-based artificial neural networks QSAR for ligand biological activity predictions},
  author={Myint, Kyaw-Zeyar and Wang, Lirong and Tong, Qin and Xie, Xiang-Qun},
  journal={Molecular pharmaceutics},
  volume={9},
  number={10},
  pages={2912--2923},
  year={2012},
  publisher={ACS Publications}
}

@article{rogers2010extended,
  title={Extended-connectivity fingerprints},
  author={Rogers, David and Hahn, Mathew},
  journal={Journal of chemical information and modeling},
  volume={50},
  number={5},
  pages={742--754},
  year={2010},
  publisher={ACS Publications}
}

@article{kingma2014adam,
  title={Adam: A method for stochastic optimization},
  author={Kingma, Diederik P and Ba, Jimmy},
  journal={arXiv preprint arXiv:1412.6980},
  year={2014}
}

@article{imakura2022dc,
  title={DC-COX: Data collaboration Cox proportional hazards model for privacy-preserving survival analysis on multiple parties},
  author={Imakura, Akira and Tsunoda, Ryoya and Kagawa, Rina and Yamagata, Kunihiro and Sakurai, Tetsuya},
  journal={Journal of Biomedical Informatics},
  pages={104264},
  year={2022},
  publisher={Elsevier}
}

@article{pearson1901liii,
  title={LIII. On lines and planes of closest fit to systems of points in space},
  author={Pearson, Karl},
  journal={The London, Edinburgh, and Dublin philosophical magazine and journal of science},
  volume={2},
  number={11},
  pages={559--572},
  year={1901},
  publisher={Taylor \& Francis}
}

@article{barker2003partial,
  title={Partial least squares for discrimination},
  author={Barker, Matthew and Rayens, William},
  journal={Journal of Chemometrics: A Journal of the Chemometrics Society},
  volume={17},
  number={3},
  pages={166--173},
  year={2003},
  publisher={Wiley Online Library}
}

@article{breiman2001random,
  title={Random forests},
  author={Breiman, Leo},
  journal={Machine learning},
  volume={45},
  pages={5--32},
  year={2001},
  publisher={Springer}
}

@article{he2003locality,
  title={Locality preserving projections},
  author={He, Xiaofei and Niyogi, Partha},
  journal={Advances in neural information processing systems},
  volume={16},
  year={2003}
}

@article{mcinnes2018umap,
  title={Umap: Uniform manifold approximation and projection for dimension reduction},
  author={McInnes, Leland and Healy, John and Melville, James},
  journal={arXiv preprint arXiv:1802.03426},
  year={2018}
}

@inproceedings{mcmahan2017communication,
  title={Communication-efficient learning of deep networks from decentralized data},
  author={McMahan, Brendan and Moore, Eider and Ramage, Daniel and Hampson, Seth and y Arcas, Blaise Aguera},
  booktitle={Artificial intelligence and statistics},
  pages={1273--1282},
  year={2017},
  organization={PMLR}
}

@article{konevcny2016federated,
  title={Federated optimization: Distributed machine learning for on-device intelligence},
  author={Kone{\v{c}}n{\`y}, Jakub and McMahan, H Brendan and Ramage, Daniel and Richt{\'a}rik, Peter},
  journal={arXiv preprint arXiv:1610.02527},
  year={2016}
}

@article{hossen2022federated,
  title={Federated machine learning for detection of skin diseases and enhancement of internet of medical things (IoMT) security},
  author={Hossen, Md Nazmul and Panneerselvam, Vijayakumari and Koundal, Deepika and Ahmed, Kawsar and Bui, Francis M and Ibrahim, Sobhy M},
  journal={IEEE journal of biomedical and health informatics},
  year={2022},
  publisher={IEEE}
}

@article{hard2018federated,
  title={Federated learning for mobile keyboard prediction},
  author={Hard, Andrew and Rao, Kanishka and Mathews, Rajiv and Ramaswamy, Swaroop and Beaufays, Fran{\c{c}}oise and Augenstein, Sean and Eichner, Hubert and Kiddon, Chlo{\'e} and Ramage, Daniel},
  journal={arXiv preprint arXiv:1811.03604},
  year={2018}
}

@article{imakura2023non,
  title={Non-readily identifiable data collaboration analysis for multiple datasets including personal information},
  author={Imakura, Akira and Sakurai, Tetsuya and Okada, Yukihiko and Fujii, Tomoya and Sakamoto, Teppei and Abe, Hiroyuki},
  journal={Information Fusion},
  pages={101826},
  year={2023},
  publisher={Elsevier}
}

@article{imakura2023another,
  title={Another use of SMOTE for interpretable data collaboration analysis},
  author={Imakura, Akira and Kihira, Masateru and Okada, Yukihiko and Sakurai, Tetsuya},
  journal={Expert Systems with Applications},
  pages={120385},
  year={2023},
  publisher={Elsevier}
}

@article{imakura2021accuracy,
  title={Accuracy and privacy evaluations of collaborative data analysis},
  author={Imakura, Akira and Bogdanova, Anna and Yamazoe, Takaya and Omote, Kazumasa and Sakurai, Tetsuya},
  journal={In: The Second AAAI Workshop on Privacy-Preserving Artificial Intelligence (PPAI-21)},
  year={2021}
}

\end{document}